\documentclass[submission,copyright,creativecommons]{eptcs}
\providecommand{\event}{M4M 2017} 
\usepackage{breakurl}             
\usepackage{underscore}           
\usepackage{amsmath, amsthm}
\usepackage{graphicx}
\usepackage{amssymb}

\title{A Gentle Introduction to Epistemic Planning: \\ The DEL Approach}
\author{Thomas Bolander 
\institute{DTU Compute \\ Technical University of Denmark \\ Copenhagen, Denmark}
\email{tobo@dtu.dk}
}
\def\titlerunning{A Gentle Introduction to Epistemic Planning: The DEL Approach}
\def\authorrunning{Thomas Bolander}

\usepackage{color, url}
\usepackage{pgf, tikz}
\usepackage{enumerate}
\usepackage{paralist}
\usepackage{subcaption}

\newcommand{\agents}{\mathcal{A}}
\newcommand{\model}{\mathcal{M}}

\def\lang{\mathcal{L}_{\textnormal{KC}}} 

\theoremstyle{definition}
\newtheorem{definition}{Definition}
\newtheorem{example}{Example}

\theoremstyle{plain}

\def\dom{\operatorname{Dom}}

\def\globals{\textit{Globals}}

\newcommand{\sglobal}{S^\text{gl}}

\let\phi\varphi

\usetikzlibrary{trees,arrows,arrows,shapes,positioning,calc}
\tikzset{
  uworld/.style={circle, fill, inner sep=1.5pt, outer sep=2pt, anchor=base},
  udworld/.style={circle, draw, inner sep=2.7pt, outer sep=2pt, anchor=base}}

\newcommand{\leaveout}[1]{}

\DeclareMathAlphabet{\mathcal}{OMS}{cmsy}{m}{n}

\begin{document}
\maketitle

\begin{abstract}
Epistemic planning can be used for decision making in multi-agent situations
with distributed knowledge and capabilities. Dynamic Epistemic Logic (DEL) has been shown to provide a very natural and expressive framework for epistemic planning. In this paper, we aim to give an accessible introduction to DEL-based epistemic planning. The paper starts with the most classical framework for planning, STRIPS, and then moves towards epistemic planning in a number of smaller steps, where each step is motivated by the need to be able to model more complex planning scenarios. 
\end{abstract}

\section{Introduction}

Automated planning is a branch of artificial intelligence concerned with computing plans (sequences of actions) leading to some desired goal. A human or robot could e.g. have the goal of picking up a parcel at the post office, and then the problem becomes to find a succesful sequence of actions achieving this. Epistemic planning is the enrichment of planning with epistemic notions, that is, knowledge and beliefs. The human or robot might have to reason about epistemic aspects such as: Do I know at which post office the parcel is? If not, who would be relevant to ask? Maybe the parcel is a birthday present for my daughter, and I want to ensure that she doesn’t get to know, and have to plan my actions accordingly (make sure she doesn't see me with the parcel). The epistemic notions are usually formalised using an epistemic logic. Epistemic planning can naturally be seen as the combination of automated planning with epistemic logic, relying on ideas, concepts and solutions from both areas.

In general, epistemic planning considers the following problem: Given my current state of knowledge, and a desirable state of knowledge, how do I get from one to the other? It is of central importance in settings where agents need to be able to reason about their own lack of knowledge, and e.g. make plans of how to achieve the required knowledge. It is also essential in multi-agent planning, where succesful coordination and collaboration can only be expected if agents are able to reason about the knowledge, uncertainty and capabilities of the other agents.

In this gentle introduction to epistemic planning, the focus will be on the DEL approach: Using Dynamic Epistemic Logic (DEL) as the underlying formalism. We start with the most classical framework for planning, STRIPS, and then stepwise we expand and generalise the framework until finally reaching full multi-agent planning based on dynamic epistemic logic. Each of these steps will be based on the need to be able to formalise specific planning scenarios. 


In Section~\ref{sect:classical} we will first introduce classical planning domains and planning tasks. Then we move to present the basics of the STRIPS planning framework in Section~\ref{sect:stripsplanning}, and propositional planning in Section~\ref{sect:propplan}. We then slowly progress towards defining the epistemic planning framework via first defining belief states and conditional actions in Section~\ref{sect:beliefstates}, and then epistemic logic and dynamic epistemic logic in Sections~\ref{sect:epistemiclogic}--\ref{sect:dynamicepistemiclogic}. In Section~\ref{sect:epistemicplanningtasks} we finally define epistemic planning tasks, and study various extensions and generalisations in Sections~\ref{sect:conditionalepistemicplanning}--\ref{sect:multiagent}. In Section~\ref{sect:complexity} we very briefly study complexity issues of epistemic planning, and we round off with a discussion of alternative approaches to epistemic planning in Section~\ref{sect:alternative}.

\section{Classical planning domains and planning tasks}\label{sect:classical}
\begin{example}\label{ex:birthday1}
Suppose a father has his daughter's birthday coming up, and he ordered a present for her which is now at the post office. His goal is to be able to give her the present the next day. This is a \textbf{planning task} (sometimes called a \textbf{planning problem}): He needs to compute a plan to achieve the goal. In a planning task, one is given an \textbf{initial state}, a set of \textbf{goal states} and a set of available \textbf{actions}. The problem is now to compute a sequence of actions (a \textbf{plan}) that, if executed in the initial state, will lead to one of the goal states. In the birthday present example, the initial state describes that the present is at the post office and not yet wrapped. The goal states are those in which the present is at home and wrapped (ready to be given on the next day). The available actions could be actions like going from home to the post office, going from the post office to home, picking up the present at the post office, and wrapping the present. Of course we do not need to limit ourselves to only allowing these specific actions, but could have general actions for going from a location $A$ to a location $B$, general actions for picking up an object at a location, and general actions for wrapping an object that you are currently holding. 
\end{example}
To allow us to reason formally about planning tasks and plans, and to allow computers and robots to compute plans, we need an appropriate formalism to describe such objects. The simplest approach is to define planning tasks in terms of state-transition systems. 
\begin{definition}{\cite{ghal.ea:auto}}
A \textbf{(restricted) state-transition system} (also called a \textbf{classical planning domain} or simply a \textbf{state space}) is $\Sigma = (S,A,\gamma)$ where:
\begin{itemize}
  \item $S$ is a finite or recursively enumerable set of \textbf{states}.
  \item $A$ is a finite or recursively enumerable set of \textbf{actions}.
  \item $\gamma: S \times A \hookrightarrow S$ is a computable partial \textbf{state-transition function}.
\end{itemize}
When $\gamma(s,a)$ is defined, $a$ is said to be \textbf{applicable} in $s$. When $\pi = a_1; \cdots; a_n$ is a sequence of actions from $A$, we write $\gamma(s,\pi)$ for $\gamma(\dots \gamma(\gamma( \gamma(s,a_1), a_2), a_3), \dots, a_n)$.\footnote{So $\gamma(s,\pi)$ is the result of executing the actions of $\pi$ in sequence starting in $s$.} 
\end{definition}

\begin{definition}{\cite{ghal.ea:auto}}
A \textbf{classical planning task} is a triple $(\Sigma,s_0,S_g)$ where:
\begin{itemize}
  \item $\Sigma = (S,A,\gamma)$ is a state-transition system (a classical planning domain).
  \item $s_0 \in S$ is the \textbf{initial state}.
  \item $S_g \subseteq S$ is the set of \textbf{goal states}. 
\end{itemize}
A \textbf{solution} to a classical planning task $((S,A, \gamma),s_0,S_g)$ is a finite sequence of actions (a \textbf{plan}) $\pi = a_1; \cdots; a_n$ from $A$ such that $\gamma(s_0,\pi) \in S_g$. The \textbf{length} of a solution $\pi$ is the number of actions in $\pi$. 
\end{definition}
\begin{example}
\begin{figure}
\begin{tikzpicture}[every node/.style={auto},>=stealth]
\node[circle,fill,inner sep=1.3pt,label=below:$s_1$] (s1) at (0,0) {};
\path[->] (s1) edge [loop above] node {\scriptsize go home} ();
\node[circle,fill,inner sep=1.3pt,label=below:$s_2$] (s2) at (3,0) {};
\path[->,very thick] (s1) edge[bend right] node[below] {\scriptsize go to post office} (s2);
\path[->] (s2) edge[bend right] node[above] {\scriptsize go home} (s1);
\node[circle,fill,inner sep=1.3pt,label=below:$s_3$] (s3) at (6,0) {};
\path[->,very thick] (s2) edge node[below] {\scriptsize pick up present} (s3);
\path[->] (s3) edge [loop above] node {\scriptsize go to post office} ();
\node[circle,fill,inner sep=1.3pt,label=below:$s_4$] (s4) at (10,0) {};
\path[->] (s4) edge [loop above] node {\scriptsize go home} ();
\path[->,very thick] (s3) edge[bend right] node[below] {\scriptsize go home} (s4);
\path[->] (s4) edge[bend right] node[above] {\scriptsize go to post office} (s3);
\node[circle,fill,inner sep=1.3pt,label=below:$s_5$] (s5) at (13,0) {};
\path[->] (s5) edge [loop above] node {\scriptsize go home} ();
\path[->,very thick] (s4) edge node[below] {\scriptsize wrap present} (s5);
\node[circle,fill,inner sep=1.3pt,label=below:$s_6$] (s6) at (10,-2) {};
\path[->] (s3) edge[bend right=15] node[below,xshift=-6mm] {\scriptsize wrap present} (s6);
\path[->] (s6) edge [loop above] node {\scriptsize go to post office} ();
\path[->] (s6) edge[bend right=10] node[below,xshift=-3mm,yshift=6mm] {\scriptsize go home} (s5);
\path[->] (s5) edge[out=-30,in=-30] node[below,xshift=8mm] {\scriptsize go to post office} (s6);
\end{tikzpicture}
\caption{A state-transition system for the birthday present example.}\label{figu:statetrans}
\end{figure}
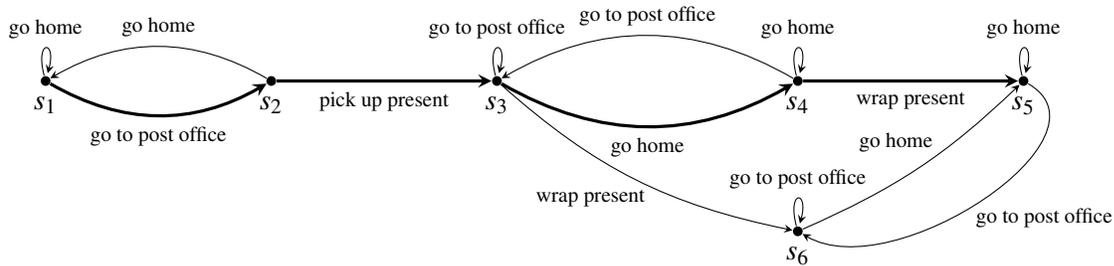
Consider the birthday present example from Example~\ref{ex:birthday1}. The available actions and their corresponding state transitions could be presented by the state-transitions system of Figure~\ref{figu:statetrans}. Here we have $\Sigma = (S,A,\gamma)$ with $S=\{s_1,s_2,s_3,s_4,s_5,s_6 \}$, $A= \{\text{go to post office},$ $\text{go home},$ $\text{pick up present},$ $\text{wrap present} \}$, and $\gamma$ is as given in the figure (e.g. $\gamma(s_1, \text{go to post office}) = s_2$ since there is an edge labelled ``go to post office'' from $s_1$ to $s_2$). Note that ``wrap present'' is only applicable after ``pick up present'' has been executed: it is necessary to get hold of the present before it can be wrapped. The planning task of Example~\ref{ex:birthday1} can then be represented as the classical planning task $(\Sigma,s_0,S_g)$ where $s_0 = s_1$ and $S_g = \{s_5\}$. A solution is highlighted in Figure~\ref{figu:statetrans}:
\[
   \pi = \text{go to post office}; \text{pick up present}; \text{go home}; \text{wrap present}.
\]
\end{example}
Representing planning tasks directly as state-transition systems has some important weaknesses: 1)~there is no internal structure on states and actions to understand or clarify what they represent; 2)~state-transition systems are normally of exponential size in the number of propositional variables required to describe the states. To exemplify the second weakness, if the planning task was to take home $n$ parcels from the post office, the state space (state-transition system) would be of size $\geq 2^n$: each parcel could either be at the post office or at home. This is so even though the length of the shortest solution  would only be linear in $n$ (e.g.\ bring the parcels home one at a time). To address both of these weaknesses one introduces logical structures on states and actions. This is e.g.\ done in the planning language STRIPS to be introduced next.

\section{STRIPS planning}\label{sect:stripsplanning}
The classical language for describing states and actions in the field of automated planning is STRIPS~\cite{fike.ea:stri}. We will here introduce STRIPS slightly informally, and the interested reader is referred to~\cite{russ.ea:arti} for more details. The reader familiar with STRIPS planning can skip this section, but might want to briefly look at the examples. 

\newcommand{\At}{\textsf{At}}
\newcommand{\Me}{\textsf{Me}}
\newcommand{\Home}{\textsf{Home}}
\newcommand{\Present}{\textsf{Present}}
\newcommand{\PostOffice}{\textsf{PostOffice}}
\newcommand{\IsAgent}{\textsf{IsAgent}}
\newcommand{\IsLocation}{\textsf{IsLocation}}
\newcommand{\IsObject}{\textsf{IsObject}}
\newcommand{\Father}{\textsf{Father}}
\newcommand{\Go}{\textsf{Go}}
\newcommand{\PickUp}{\textsf{PickUp}}
\newcommand{\agt}{\textit{agt}}
\newcommand{\from}{\textit{from}}
\newcommand{\myto}{\textit{to}}
\newcommand{\obj}{\textit{obj}}
\newcommand{\Has}{\textsf{Has}}
\newcommand{\Wrap}{\textsf{Wrap}}
\newcommand{\Wrapped}{\textsf{Wrapped}}
\begin{figure}
\[ \small
  \begin{array}{l}
  \textsc{Action}: \Go(\agt,\from,\myto) \\
  \textsc{Precond}: \At(\agt,\from) \land \IsAgent(\agt) \land \IsLocation(\from) \land \IsLocation(\myto) \\
  \textsc{Effect}: \At(\agt,\myto) \land \neg \At(\agt,\from) \\ \\
  \textsc{Action}: \PickUp(\agt,\obj,\from) \\
  \textsc{Precond}: \At(\agt,\from) \land \At(\obj,\from) \land \neg \Has(\agt,\obj) \land \IsAgent(\agt) \land \IsObject(\obj) \land \IsLocation(\from) \\
  \textsc{Effect}: \Has(\agt,\obj) \land \neg \At(\obj,\from) \\ \\
  \textsc{Action}: \Wrap(\agt,\obj) \\
  \textsc{Precond}: \Has(\agt,\obj) \land \neg \Wrapped(\obj) \land \IsAgent(\agt) \land \IsObject(\obj) \\
  \textsc{Effect}: \Wrapped(\obj) \\ \\ 
\end{array}
\]
\caption{The set of STRIPS action schemas for the birthday present example.}\label{figu:stripsactions}
\end{figure}
In STRIPS, states are represented as sets of ground atoms of a function-free first-order language $\mathcal{L}$. In the birthday present example, the initial state could e.g.\ be described by
\begin{equation}\label{form:firsts0}
  \begin{array}{rl}
  s_0 = &\{ \At(\Father,\Home), \At(\Present,\PostOffice), \IsAgent(\Father), \IsLocation(\Home),  \\    
          &\IsLocation(\PostOffice), \IsObject(\Present) \}
        \end{array}
\end{equation}
where $\At(x,y)$ is a predicate for expressing that object (or agent) $x$ is at location $y$. The predicates $\IsAgent(x)$, $\IsLocation(x)$ and $\IsObject(x)$ are true of agents, objects and locations, respectively (alternatively, one could use a sorted first-order language and then omit these predicates). Actions are described via so-called \textbf{action schemas}. The actions of \emph{going from one location to another}, \emph{picking up an object} and \emph{wrapping an object} can be expressed by the STRIPS action schemas provided in Figure~\ref{figu:stripsactions}, where predicates have names starting with upper-case letters, and variables have names starting with lower-case letters.
Each action schema has a \textbf{name}, a \textbf{precondition} and an \textbf{effect}. Preconditions and effects are conjunctions of literals of the first-order language $\mathcal{L}$. A \textbf{ground action} is achieved by instantiating all variables of an action schema with constants of $\mathcal{L}$. For instance, 
\[ \small
  \begin{array}{l}
  \textsc{Action}: \Go(\Father,\Home,\PostOffice) \\
  \textsc{Precond}: \At(\Father,\Home) \land \IsAgent(\Father) \land \IsLocation(\Home) \land \IsLocation(\PostOffice) \\
  \textsc{Effect}: \At(\Father,\PostOffice) \land \neg \At(\Father,\Home) \\ \\
\end{array}
\]
The precondition of a ground action describes what has to be true for the action to be applicable.\footnote{More formally, a ground action $a$ is applicable in a state $s$ if $s \models \textsc{Precond}(a)$ where $\models$ denotes semantical entailment on propositional logic over the set of ground atoms of $\mathcal{L}$.} It is easy to check that $\Go(\Father,\Home,\PostOffice)$ is applicable in the initial state $s_0$ defined by (\ref{form:firsts0}) (all precondition atoms of the action occur in $s_0$). The effect of a ground action describes how the state is modified when the action is executed. The effect of $\Go(\Father,\Home,\PostOffice)$ expresses that $\At(\Father,\PostOffice)$ becomes true, and that $\At(\Father,\Home)$ becomes false. Hence the result of executing $\Go(\Father,\Home,\PostOffice)$ in $s_0$ will be the state
\[
  \begin{array}{rl}
  s_1 = &\{ \At(\Father,\PostOffice), \At(\Present,\PostOffice), \IsAgent(\Father), \IsLocation(\Home),  \\    
          &\IsLocation(\PostOffice), \IsObject(\Present) \}
        \end{array}
\]

Any finite set of STRIPS action schemas $A$ \textbf{induce} a state-transition system (classical planning domain) $\Sigma = (S,A',\gamma)$ by:
\begin{itemize}
  \item $S = 2^P$, where $P$ is the set of ground atoms of $\mathcal{L}$.
  \item $A' = \{ \text{all ground instances of the action schemas in $A$} \}$
  \item $\gamma(s,a) = \begin{cases} 
    (s - \{ \phi \mid \neg \phi \text{ is a negative literal of } \textsc{Effect}(a))\ \cup \\
    \quad \{ \phi \mid \phi \text{ is a positive literal of }\textsc{Effect}(a) \} &\text{if $s \models \textsc{Precond}(a)$} \\
    \text{undefined} &\text{otherwise}
    \end{cases}$
\end{itemize}
\newcommand{\Hom}{\textsf{H}}
\newcommand{\Pre}{\textsf{P}}
\newcommand{\PO}{\textsf{PO}}
\newcommand{\F}{\textsf{F}}
The state-transition system induced by the STRIPS schemas of Figure~\ref{figu:stripsactions} is provided in Figure~\ref{figu:statetrans2}. The \textbf{rigid atoms}, those that cannot change truth-value, have been omitted. Furthermore, each constant name is abbreviated by the capital letters it contains (so for instance \PostOffice\ is abbreviated \PO).
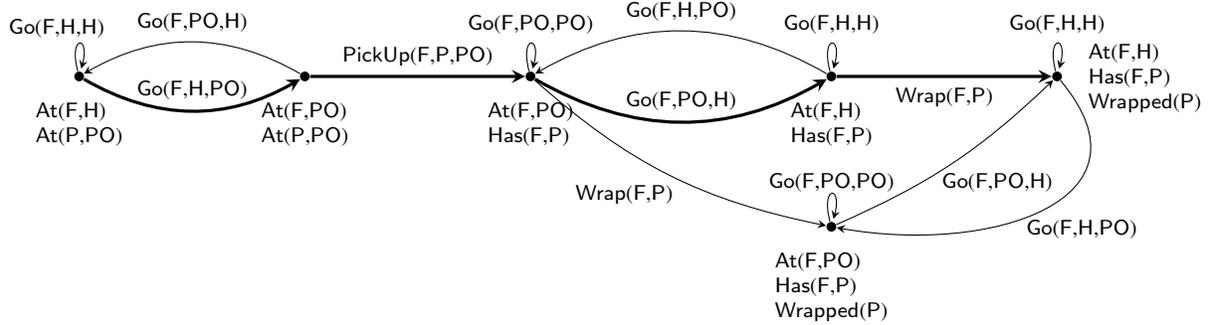
\begin{figure}
\begin{tikzpicture}[every node/.style={auto},>=stealth]
\node[circle,fill,align=left,inner sep=1.3pt,label=below:{\begin{tabular}{l} \scriptsize \At(\F,\Hom) \\[-1.5mm] \scriptsize \At(\Pre,\PO) \end{tabular}}] (s1) at (0,0) {};
\path[->] (s1) edge [loop above] node[xshift=-3mm,yshift=-0.5mm] {\scriptsize \Go(\F,\Hom,\Hom)} ();
\node[circle,fill,inner sep=1.3pt,label=below:{\begin{tabular}{l} \scriptsize \At(\F,\PO) \\[-1.5mm] \scriptsize \At(\Pre,\PO) \end{tabular}}] (s2) at (3,0) {};
\path[->,very thick] (s1) edge[bend right] node[above] {\scriptsize \Go(\F,\Hom,\PO)} (s2);
\path[->] (s2) edge[bend right] node[above] {\scriptsize \Go(\F,\PO,\Hom)} (s1);
\node[circle,fill,inner sep=1.3pt,label=below:{\begin{tabular}{l} \scriptsize \At(\F,\PO) \\[-1.5mm] \scriptsize\Has(\F,\Pre) \end{tabular}}] (s3) at (6,0) {};
\path[->,very thick] (s2) edge node[above] {\scriptsize \PickUp(\F,\Pre,\PO)} (s3);
\path[->] (s3) edge [loop above] node {\scriptsize \Go(\F,\PO,\PO)} ();
\node[circle,fill,inner sep=1.3pt,label=below:{\begin{tabular}{l} \scriptsize \At(\F,\Hom) \\[-1.5mm] \scriptsize\Has(\F,\Pre) \end{tabular}}] (s4) at (10,0) {};
\path[->] (s4) edge [loop above] node {\scriptsize \Go(\F,\Hom,\Hom)} ();
\path[->,very thick] (s3) edge[bend right] node[above] {\scriptsize \Go(\F,\PO,\Hom)} (s4);
\path[->] (s4) edge[bend right] node[above] {\scriptsize \Go(\F,\Hom,\PO)} (s3);
\node[circle,fill,inner sep=1.3pt,label=right:{\begin{tabular}{l} \scriptsize \At(\F,\Hom) \\[-1.5mm] \scriptsize\Has(\F,\Pre) \\[-1.5mm] \scriptsize\Wrapped(\Pre) \end{tabular}}] (s5) at (13,0) {};
\path[->] (s5) edge [loop above] node {\scriptsize \Go(\F,\Hom,\Hom)} ();
\path[->,very thick] (s4) edge node[below] {\scriptsize \Wrap(\F,\Pre)} (s5);
\node[circle,fill,inner sep=1.3pt,label=below:{\begin{tabular}{l} \scriptsize \At(\F,\PO) \\[-1.5mm] \scriptsize\Has(\F,\Pre) \\[-1.5mm] \scriptsize\Wrapped(\Pre) \end{tabular}}] (s6) at (10,-2) {};
\path[->] (s3) edge[bend right=15] node[below,xshift=-6mm] {\scriptsize \Wrap(\F,\Pre)} (s6);
\path[->] (s6) edge [loop above] node[yshift=-1mm] {\scriptsize \Go(\F,\PO,\PO)} ();
\path[->] (s6) edge[bend right=10] node[below,xshift=6mm] {\scriptsize \Go(\F,\PO,\Hom)} (s5);
\path[->] (s5) edge[out=-50,in=-10,looseness=1.5] node[below,xshift=5mm] {\scriptsize \Go(\F,\Hom,\PO)} (s6);
\end{tikzpicture}
\caption{The state-transition system induced by the action schemas of Figure~\ref{figu:stripsactions}.}\label{figu:statetrans2}
\end{figure}
Note that the resulting state-transition system is isomorphic to the state-transition system of Figure~\ref{figu:statetrans}. 

The advantage of the current STRIPS representation over the previous pure state-transition representation is that there is now structure on states and actions. This for instance means that the current formulation is easily generalisable, e.g.\ we can add $n-1$ additional parcels to the initial state without any need to modify the underlying action schemas. The induced state-transition system would become exponentially bigger, as earlier noted, but the size of the STRIPS action schemas would stay the same (though the size of the description of the initial state would grow linearly with $n$). 

A \textbf{STRIPS planning task} on a function-free first-order language $\mathcal{L}$ is $(A,s_0,\phi_g)$ where $A$ is a set of STRIPS action schemas over $\mathcal{L}$, $s_0$ is a set of ground atoms over $\mathcal{L}$ and $\phi_g$ is a conjunction of ground literals over $\mathcal{L}$ called the \textbf{goal formula}. Any STRIPS planning task $(A,s_0,\phi_g)$ induces a classical planning task $(\Sigma,s_0,S_g)$ by letting $\Sigma$ be the state-transition system induced by $A$ and letting $S_g = \{s \in S \mid s \models \phi_g \}$. A \textbf{solution} to a STRIPS planning task is then a solution to the induced classical planning task. 

\begin{example}\label{ex:stripsplanningtask}
The birthday present example can be represented as a STRIPS planning task $(A,s_0,\phi_g)$ where $A$ is the set of action schemas of Figure~\ref{figu:stripsactions}, $s_0$ is defined by (\ref{form:firsts0}), and $\phi_g = \At(\Father,\Home) \land \Has(\Father,\Present) \land \Wrapped(\Present)$. This is a STRIPS planning task on the function-free first order language $\mathcal{L}$ with binary predicate symbols $\At$ and $\Has$, unary predicate symbols $\Wrapped$, $\IsAgent$, $\IsLocation$ and $\IsObject$, and constant symbols $\Home$, $\PostOffice$, $\Father$, $\Present$. Consulting Figure~\ref{figu:statetrans2}, it is easy to show that a solution to this planning task is
\[
 \begin{array}{rl}
  \pi = &\Go(\Father,\Home,\PostOffice); \PickUp(\Father,\Present,\PostOffice); \\
  &\Go(\Father,\PostOffice,\Home); \Wrap(\Father,\Present).
  \end{array}
\]
\end{example}

In the field of automated planning, actions are always described compactly in an action description language like STRIPS (or e.g.\ PDDL, ADL or SAS$^+$). A lot of research effort goes into finding ways to automatically derive efficient heuristics from action schemas, such that solutions can be found with minimal search. If given an induced state-transition system of a set of STRIPS action schemas, finding a solution to a planning task becomes a simple graph search problem (find a path from $s_0$ to a state in $S_g$). This can be done in linear time in the size of the state-transition system. However, as earlier noted, the induced state-transition system is often exponential in the size of the action schemas. Hence the complexity of computing solutions or deciding whether a solution exists (the \textbf{plan existence problem}) is in automated planning always measured in the size of the compact action schema representation. This is different from many formalisms in logic that consider plans or strategies and where complexity is measured in terms of the size of the state space. This is e.g.\ why epistemic planning based on ATEL in \cite{hoek.ea:trac} can be claimed to be tractable, even though already basic propositional STRIPS planning, which is much less expressive, is intractable. For most planning domains considered in automated planning (e.g.\ the planning domains of the International Planning Competition, IPC), calculating the entire state-transition system is computationally infeasible, and the goal is then for the heuristics to be sufficiently efficient that only the most relevant parts of the state-transition system are explored.

\section{Propositional planning tasks}\label{sect:propplan}
Even though STRIPS action schemas are formulated using first-order logic, for most purposes we can consider STRIPS as a planning formalism based on propositional logic. To see this, we first define \emph{propositional planning tasks}.
\begin{definition}
A \textbf{propositional planning task}\footnote{Called a \textbf{set-theoretic planning task} in \cite{ghal.ea:auto}.} on a finite set of atomic propositions $P$ is $(A,s_0,\phi_g)$ where
\begin{itemize}
  \item $A$ is a finite set of \textbf{actions}. Each action is a pair $a = \langle pre(a), post(a) \rangle$ where $pre(a)$ and $post(a)$ are conjunctions of propositional literals over $P$. The element $pre(a)$ is called the \textbf{precondition} of $a$ and $post(a)$ its \textbf{postcondition}. 
  \item $s_0$ is the \textbf{initial state}, a propositional state over $P$ (a subset of $P$).\footnote{In general, we will identify propositional states with subsets of $P$. A subset $S \subseteq P$ represents the propositional state $s$ in which the elements of $S$ are the atomic propositions true in $s$.} 
  \item $\phi_g$ is the \textbf{goal formula}, a propositional formula over $P$.
\end{itemize}
\end{definition}
A propositional planning task $(A,s_0,\phi_g)$ on $P$ \textbf{induces} a classical planning task $((S,A,\gamma),s_0,S_g)$ in the expected way (compare with the classical planning task induced by a STRIPS planning task defined in Section~\ref{sect:stripsplanning}):
\begin{itemize}
  \item $S = 2^P$
   \item $\gamma(s,a) = \begin{cases} 
    (s - \{ p \mid \neg p \text{ is a negative literal of } post(a))\ \cup \\
    \quad \{ p \mid p \text{ is a positive literal of } post(a) \} &\text{if $s \models pre(a)$} \\
    \text{undefined} &\text{otherwise}
    \end{cases}$
  \item $S_g = \{ s \in S \mid s \models \phi_g \}$. 
\end{itemize}
A \textbf{solution} to a propositional planning task is any solution to the induced classical planning task. 

For any function-free first-order language $\mathcal{L}$, let $P_\mathcal{L}$ denote the set of ground atoms of $\mathcal{L}$. The set $P_\mathcal{L}$ can be thought of as the atomic propositions of a propositional language. Any quantifier-free ground formula of $\mathcal{L}$ is then at the same time a formula of propositional logic over $P_\mathcal{L}$. Any STRIPS planning task $(A,s_0,\phi_g)$ on $\mathcal{L}$ \textbf{induces} a propositional planning task $(A',s_0,\phi_g)$ on $P_\mathcal{L}$ by simply letting
\[
  A' = \{ \langle \textsf{Precond}(a), \textsf{Effect}(a) \rangle \mid a \text{ is a ground instance of an action schema in } A \}.
\]
It is easy to show that the STRIPS planning task $(A,s_0,\phi_g)$ and its induced propositional planning task $(A',s_0,\phi_g)$ both induce the same classical planning task. They thus also have the same solutions. Note that the conventions for preconditions and effects are a bit different in propositional planning tasks than in STRIPS planning tasks. We now write precondition and effect pairs of an action $a$ as a pair of the form $\langle pre(a), post(a) \rangle$, where we have also relabelled effects as \emph{postconditions}. The point of both these changes is to gradually move away from the classical conventions of STRIPS planning into the conventions of dynamic epistemic logic that we will later present a planning framework based on.
\newcommand{\Agt}{\textit{Agt}}
\newcommand{\Loc}{\textit{Loc}}
\newcommand{\Obj}{\textit{Obj}}
\begin{example}\label{ex:propplantask}
The birthday present example can be represented as the propositional planning task $(A,s_0,\phi_g)$ below. It is a simplified version of the propositional planning task induced by the STRIPS planning task of Example~\ref{ex:stripsplanningtask}, where we have done away with the rigid atoms that are no longer necessary. In the definitions below, $\Agt$ is a set of agent names (including \Father), $\Loc$ is a set of locations (including \Home\ and \PostOffice) and $\Obj$ is a set of objects (including \Present).
\begin{itemize}
  \item $A = \{ \Go(\agt,\from,\myto) \mid \agt \in \Agt\ \&\ \from, \myto \in \Loc \} \cup \{ \PickUp(\agt,\obj,\from) \mid \agt \in \Agt\ \&\  \obj \in \Obj\ \&\ \from \in \Loc \} \cup \{ \Wrap(\agt,\obj) \mid \agt \in \Agt\ \& \obj \in \Obj \}$ where, for all $\agt \in \Agt$, all $\from, \myto \in \Loc$ and all $\obj \in \Obj$,
  \begin{itemize}
    \item $\Go(\agt,\from,\myto) = \langle \At(\agt,\from), \At(\agt,\myto) \land \neg \At(\agt,\from) \rangle$
    \item $\PickUp(\agt,\obj,\from) = \langle \At(\agt,\from) \land \At(\obj,\from) \land \neg \Has(\agt,\obj), 
        \Has(\agt,\obj) \rangle$
     \item $\Wrap(\agt,\obj) = \langle \Has(\agt,\obj) \land \neg \Wrapped(\obj), \Wrapped(\obj) \rangle$.
  \end{itemize} 
  \item $s_0 = \{ \At(\Father,\Home), \At(\Present,\PostOffice) \}$.
  \item $\phi_g = \At(\Father,\Home) \land \Has(\Father,\Present) \land \Wrapped(\Present)$.
\end{itemize}
Note that expressions like $\At(\agt,\from)$ with $\agt \in \Agt$ and $\from \in \Loc$ are no longer considered as ground atoms of the original first-order language $\mathcal{L}$, but as atoms of propositional logic over $P_\mathcal{L}$. A solution to this planning task is exactly as to the original STRIPS version: $\pi = \Go(\Father,\Home,\PostOffice);$ $\PickUp(\Father,\Present,\PostOffice);$  $\Go(\Father,\PostOffice,\Home);$ $\Wrap(\Father,\Present)$.
\end{example}
Since any STRIPS planning task can be \emph{propositionalised} as above, it means we can now work in a simpler formalism, propositional logic, which also makes it easier to generalise the formalism to e.g.\ planning with partial observability, non-determinism or epistemic planning.\footnote{In certain cases the difference between expressing a planning task in the \emph{lifted} first-order STRIPS representation and its induced propositionalisation becomes essential: for some of the complexity results measured in the size of the planning tasks; for practical convenience of representation; or e.g.\ for learning actions/action models~\cite{Walsh:2008fk,DBLP:conf/uai/MouraoZPS12}. However, for the purposes of this paper, the grounded/propositionalised representation is sufficient.}

\section{Belief states, partial observability, and conditional actions}\label{sect:beliefstates}
\newcommand{\PostOfficeOne}{\textsf{PostOffice1}}
\newcommand{\PostOfficeTwo}{\textsf{PostOffice2}}
 Consider the birthday present example formalised as the propositional planning task of Example~\ref{ex:propplantask}. Assume now that there is not only one, but two, local post offices, and the father does not know in which one the parcel is. We can assume $\Loc$ correspondingly contains two constants, $\PostOfficeOne$ and $\PostOfficeTwo$. To represent this modified planning task we need two changes in the underlying formalism: \emph{belief states} and \emph{conditional actions}. We need belief states to represent the uncertainty of the agent. A \textbf{belief state} in this setting is a set of propositional states, that is,  a set of subsets of $P$ (where $P$ is the set of atomic propositions). The initial belief state of our agent is now:
 \begin{equation}\label{eq:initbeliefstate} \small
 s_0 = \{ \{ \At(\Father,\Home), \At(\Present,\PostOfficeOne) \}, \{ \At(\Father,\Home), \At(\Present,\PostOfficeTwo) \} \}.
 \end{equation}
 The state $s_0$ contains two propositional states, where the first one represents the situation where the present is in \PostOfficeOne, and the second presents the situation where it is in \PostOfficeTwo.
 We define a propositional formula $\phi$ to be \textbf{true} in a belief state $s$, written $s \models \phi$, if $\phi$ is true in all propositional states of $s$. Hence we have
 \begin{enumerate}
    \item[(1)] $s_0 \models \At(\Father,\Home)$
    \item[(2)] $s_0 \not\models \At(\Present,\PostOfficeOne)$
    \item[(3)] $s_0 \not\models \At(\Present,\PostOfficeTwo)$
    \item[(4)] $s_0 \models \At(\Present,\PostOfficeOne) \lor \At(\Present,\PostOfficeTwo)$.
 \end{enumerate}
 This represents the \textbf{internal perspective} of the father in the belief state $s_0$: He can verify (knows) that he is home (1) and can verify (knows) that the present is in \PostOfficeOne\ or \PostOfficeTwo\ (4), but doesn't know which (2--3). Planning in the space of belief states rather than propositional states is called \textbf{planning under partial observability} (and planning on propositional states is then called \textbf{planning under full observability}). In the following, and in line with modal logic, we will call the elements of belief states \textbf{worlds}.  

To represent the modified example we also need to allow \emph{conditional actions}. The agent can attempt to pick up the present at either of the two post offices, but whether he is succesful is conditional on whether it is the correct one. Symmetric to the generalisation from representing states as subsets of $P$ to sets of such subsets, we can generalise actions from being pairs $\langle pre(a), post(a) \rangle$ to be sets of such pairs. We can then represent the attempted pickup action by: 
\newcommand{\TryPickUp}{\textsf{TryPickUp}}
\begin{equation}\label{eq:trypickupgeneral}
\begin{array}{l}
  \TryPickUp(\agt,\obj,\from) = \{  \\
  \quad \langle \At(\agt,\from) \land \At(\obj,\from) \land \neg \Has(\agt,\obj), 
        \Has(\agt,\obj)  \land \neg \At(\obj,\from) \rangle, \\
        \quad \langle \At(\agt,\from) \land \neg \At(\obj,\from), \top \rangle\ \ \\ \}
    \end{array}
\end{equation}
where the postcondition $\top$ means that nothing changes. From now on we will, in line with the literature on dynamic epistemic logic, call pairs $\langle pre(e), post(a) \rangle$ \textbf{events}. So a conditional action like the one above is a set of events: a set of the possible things that might happen when the action is executed. 
The \TryPickUp\ action above expresses that if the agent and the object are in the same location, the object will be successfully picked up (the first event of the action), and otherwise nothing will happen (the second event of the action). 

Given a belief state $s$ represented as a set of worlds and an action action $a$ represented as a set of events, we can define a generalised transition function by
\begin{equation}\label{form:gammabeliefstates}
  \gamma(s,a) = \{ \gamma(w,e) \mid w \in s, e \in a, w \models pre(e) \}.\footnote{This is consistent with how the transition function is defined for conditional actions in \cite{ghal.ea:auto}, but only for actions in which the events have pairwise mutually inconsistent preconditions. If two events have mutually \emph{consistent} preconditions, it means there exists states in which both are applicable. This can be interpreted in two ways. Either it represents \emph{non-determinism} where only one of the events can actually take place when the action is executed. Or it represents a situation where multiple events occur in parallel. In \cite{ghal.ea:auto}, the latter interpretation is used. In this paper and in dynamic epistemic logic, the first interpretation is used.}
\end{equation}
 \newcommand{\POne}{\textsf{PO1}}
 \newcommand{\PTwo}{\textsf{PO2}}
Thus, e.g., where we abbreviate \PostOfficeOne\ by \POne, \PostOfficeTwo\ by \PTwo\ and \Home\ by \Hom:
\[ \small
  \begin{array}{rl}
    s_1 &= \gamma(s_0, \Go(\Father,\Hom,\POne)) \\
           &= \gamma(\{ \{ \At(\Father,\Hom), \At(\Present,\POne) \}, \{ \At(\Father,\Hom), \At(\Present,\PTwo) \} \}, \Go(\Father,\Hom,\POne)) \\
           &= \{ \{ \At(\Father,\POne), \At(\Present,\POne) \}, \{ \At(\Father,\POne), \At(\Present,\PTwo) \} \} 
           \\ \\
   s_2 &= \gamma(s_1,\TryPickUp(\Father,\Present,\POne))  \\
          &= \{ \{ \At(\Father,\POne), \Has(\Father,\Present) \}, \{ \At(\Father,\POne), \At(\Present,\PTwo) \} \} 
  \\ \\
   s_3 &= \gamma(s_2,\Go(\Father,\POne,\PTwo))  \\
          &= \{ \{ \At(\Father,\PTwo), \Has(\Father,\Present) \}, \{ \At(\Father,\PTwo), \At(\Present,\PTwo) \} \}  \\ \\
             s_4 &= \gamma(s_3,\TryPickUp(\Father,\Present,\PTwo))  \\
          &= \{ \{ \At(\Father,\PTwo), \Has(\Father,\Present) \} \} 
          \end{array}
\]
\[
\small
\begin{array}{rl}          
            s_5 &= \gamma(s_4,\Go(\Father,\PTwo,\Home))  \\
          &= \{ \{ \At(\Father,\Home), \Has(\Father,\Present) \} \} \\ \\
   s_6 &= \gamma(s_5,\Wrap(\Father,\Present))  \\
          &= \{ \{ \At(\Father,\Home), \Has(\Father,\Present), \Wrapped(\Present) \} \} 
  \end{array}
\]
Note how the belief state goes down from cardinality 2 to cardinality 1 when going from $s_3$ to $s_4$. At plan time, when deliberating about the possible action sequences, the father doesn't know whether he will have the present after having visited only \PostOfficeOne, and he hence has to represent both possibilities. In $s_4$, however, he knows, even at plan time, that he will have the present, since if he didn't get it at \PostOfficeOne, he will be sure to get it at \PostOfficeTwo. The calculations above show that a solution to the modified planning task is 
\begin{equation}\label{eq:planmodified}
\begin{array}{rl}
  \pi = &\Go(\Father,\Hom,\POne);  \TryPickUp(\Father,\Present,\POne); \Go(\Father,\POne,\PTwo); 
  \\ &\TryPickUp(\Father,\Present,\PTwo); \Go(\Father,\PTwo,\Home); \Wrap(\Father,\Present).
  \end{array}
  \end{equation}

There are, unfortunately, several weaknesses in the current approach. An important weakness is that the formalism does so far not distinguish between between the kind of uncertainty the father has in $s_1$ and the kind of uncertainty he has in $s_2$. In $s_1$, the father has \textbf{run time uncertainty}~\cite{petr.ea:know} about the location of the present. This means that even at execution time, when the action $\Go(\Father,\Hom,\POne)$ has been executed in the initial state, the father still does not know which of the two worlds of $s_1$ is the actual one. In $s_2$, however, he should only have \textbf{plan time uncertainty}~\cite{petr.ea:know}. This means that when he is deliberating about the possible action sequences and their potential outcomes, he can not tell which of the two worlds of $s_2$ is going to be realised, but at execution time he \emph{will} know. This is because attempting to pick up the present at \POne\ has the side-effect of informing him whether the present is there or not. The distinction between plan time and run time uncertainty is not represented in our current formalism, since both kinds of uncertainty are modelled exactly the same: by simply having a set of worlds representing those situations that the agent cannot distinguish between (whether at plan time or at run time). 

The simplest and most common solution to this problem in the automated planning literature (see e.g. \cite{ghal.ea:auto,rintanen2006introduction}) is to treat \emph{observability} as a static partition on the set of possible worlds, so that certain possible worlds are always distinguishable and others never are. What becomes distinguishable at execution time is then hardcoded into this observability partition. This approach is however insufficient for our purposes. For instance, assume the present is initially located at \PTwo. Assume further that the father initially goes to \POne\ to attempt picking it up and then goes home again. Then afterwards he will be in exactly the same world as initially, namely the world satisfying $\At(\Father,\Home) \land \At(\Present,\PTwo)$. But he will not be in the same \emph{information state}: He learned that the present is not at \POne\ (and therefore also learned that it must be at \PTwo). To be able to represent this in an appropriate way we need to take the final step into planning based on epistemic logic.\footnote{In principle, in the single-agent case, we could alternatively consider modelling states as sets of sets of worlds (that is, elements of $2^{2^{2^P}}$) and actions a sets of sets of events. In a state $\{ \{ w_1^1, w_1^2, \dots, w_1^{n_1} \}, \{ w_2^1, w_2^2, \dots, w_2^{n_2} \}, \dots, \{ w_m^1, w_m^2, \dots, w_m^{n_m} \} \}$, two worlds $w_i^k$ and $w_j^l$ would then always be plan time indistinguishable, but only run time indistinguishable if $i = j$. This is essentially equivalent to representing states as models of single-agent epistemic logic (see next section). Since we are anyway going to generalise to multi-agent planning, we choose to move straight to epistemic logic, rather than consider this further intermediate step.}

\section{Epistemic logic}\label{sect:epistemiclogic}
Let $P$ be a finite set of atomic propositions (often we will take $P$ to be the set of ground atoms of a function-free first-order language as in the previous sections). Let $\mathcal{A}$ be a finite set of agents. The \textbf{epistemic language} on $P,\mathcal{A}$, denoted $\lang(P,\mathcal{A})$, is generated by the following BNF:
\[
  \phi ::= \top \mid \bot \mid p \mid \neg \phi \mid \phi \land \phi \mid K_i \phi \mid C \phi, 
\]
where $p \in P$ and $i \in \mathcal{A}$. We read $K_i\phi$ as ``agent $i$ knows $\phi$'' and $C \phi$ as ``it is common knowledge that $\phi$''. When $P$ and $\mathcal{A}$ are clear from the context (or no assumptions about them are made), we will write $\lang$ for $\lang(P,\mathcal{A})$. $\mathcal{L}_\textnormal{K}$ is the language $\lang$ without the $C$ modality.
\begin{definition}
  Let $P$ and $\agents$ be as above. An \textbf{epistemic model} on $P,\agents$ is $\mathcal{M} = (W,(\sim_i)_{i \in \agents},L)$ where
  \begin{itemize}
    \item The \textbf{domain} $W$ is a non-empty finite set of \textbf{worlds}.
    \item $\sim_i\ \subseteq W \times W$ is an equivalence relation called the \textbf{indistinguishability relation} for agent $i \in \agents$.\footnote{We will later consider generalising to non-equivalence relations, but for now it is sufficient ot make this simplification and only consider equivalence relations.}
    \item $L: W \to 2^P$ is a \textbf{labelling function} assigning a propositional valuation (a set of propositions) to each world. 
  \end{itemize}
  For $W_d \subseteq W$, the pair $(\model,W_d)$ is called an \textbf{epistemic state} (or simply a \textbf{state}), and the worlds of $W_d$ are called the \textbf{designated worlds}. A state is called \textbf{global} if $W_d = \{w\}$ for some world $w$ (called the \textbf{actual world}), and we then often write $(\mathcal{M},w)$ instead of $(\mathcal{M}, \{w\})$. We use $\sglobal(P,\agents)$ to denote the set of global states (or simply $\sglobal$ if $P$ and $\agents$ are clear from the context). For any state $s = (\model,W_d)$ we let $\globals(s) = \{ (\model, w) \mid w \in W_d \}$. A state $(\model,W_d)$ is called a \textbf{local state} for agent $i$ if $W_d$ is closed under $\sim_i$ (that is, if $w \in W_d$ and $w \sim_i v$ implies $v \in W_d$). Given a state $s = (\model,W_d)$, the \textbf{associated local state} of agent $i$, denoted $s^i$, is $(\model, \{ v \mid v \sim_i w \text{ and } w \in W_d \})$.
\end{definition}
\begin{definition}
Let $(\model,W_d)$ be a state on $P,\agents$ with $\model = (W, (\sim_i)_{i \in \agents}, L)$. For $i \in \agents$, $p \in P$ and $\phi, \psi \in \lang(P,\agents)$, we define truth as follows:
\[
\begin{array}{lp{5mm}cp{5mm}l}
  (\model, W_d) \models \phi && \text{iff} && (\model,w) \models \phi \text{ for all $w \in W_d$} \\
  (\model,w) \models p && \text{iff}  && p \in L(w) \\
  (\model,w) \models \neg \phi &&\text{iff} &&\model,w \not\models \phi \\
  (\model,w) \models \phi \wedge \psi &&\text{iff} &&\model,w \models \phi \text{ and } \model,w \models \psi \\
  (\model,w) \models K_i \phi &&\text{iff} &&\text{$(\model,v) \models \phi$ for all $v \sim_i w$} \\
  (\model,w) \models C \phi &&\text{iff} &&\text{$(\model,v) \models \phi$ for all $v \sim^\ast w$} 
 \end{array}\]
where $\sim^\ast$ is the transitive closure of $\bigcup_{i\in \agents} \sim_i$.
\end{definition}
Note that a formula $\phi$ is defined to be true in a non-global state $(\model,W_d)$ iff it is true in each of the global states $(\model,w)$, $w\in W_d$, it contains. This is consistent with our previous definition of truth in belief states as truth in all worlds of that state.
\begin{example}\label{ex:actionmodels}
 Let $P = P_\mathcal{L}$, where $\mathcal{L}$ is the function-free first-order language of Example~\ref{ex:stripsplanningtask} extended with constant symbols $\PostOfficeOne$ and $\PostOfficeTwo$ and let $\agents = \{ \Father \}$. Consider the initial state $s_0$ defined by (\ref{eq:initbeliefstate}). We can represent this via an epistemic model $\model = (W,\sim_{\Father},L)$ on $P,\agents$ with $W = \{w_1,w_2\}$, $\sim_{\Father}\ = W \times W$, $L(w_1) = \{ \At(\Father,\Home), \At(\Present,\PostOfficeOne) \}$ and $L(w_2) = \{ \At(\Father,\Home), \At(\Present,\PostOfficeTwo) \}$. Graphically, this epistemic model is represented as follows, using the abbreviations from earlier:
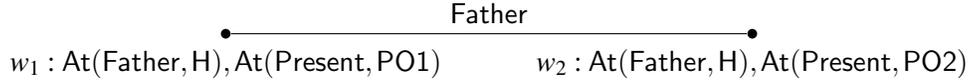
\[
  \begin{tikzpicture}[every node/.style={auto},>=stealth]
\node[circle,fill,align=left,inner sep=1.3pt,label=below:{$w_1: \At(\Father,\Hom), \At(\Present,\POne)$}] (s1) at (0,0) {};
\node[circle,fill,inner sep=1.3pt,label=below:{$w_2: \At(\Father,\Hom), \At(\Present,\PTwo)$}] (s2) at (7,0) {};
\path[-] (s1) edge node[above] {\Father} (s2);
\end{tikzpicture}
\] 
Here the nodes represent the worlds and the edges represent the indistinguishability relation (reflexive edges left out). Each node is labelled by its name and the set of atomic propositions true at the world. Consider the case where initially the present is at \PostOfficeTwo. This means that the actual world is $w_2$, and this situation can be represented by the global state $s_0 = (\model, w_2)$, that we graphically present as
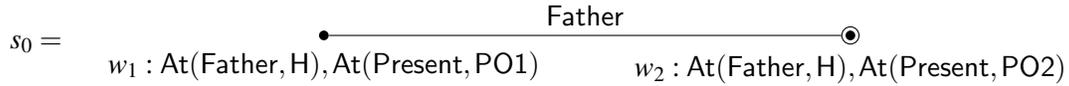
\[ \raisebox{6mm}{$s_0 =$ \quad} 
 \begin{tikzpicture}[every node/.style={auto},>=stealth]
\node[circle,fill,align=left,inner sep=1.3pt,label=below:{$w_1: \At(\Father,\Hom), \At(\Present,\POne)$}] (s1) at (0,0) {};
\node[draw,circle,inner sep=2.3pt,label=below:{$w_2: \At(\Father,\Hom), \At(\Present,\PTwo)$},after node path={node[circle,fill,inner sep=1.3pt] at (\tikzlastnode)  {}}] (s2) at (7,0) {};
\path[-] (s1) edge node[above] {\Father} (s2);
\end{tikzpicture}
\] 
We use \raisebox{0mm}{$\tikz \node[draw,circle,inner sep=2.3pt,after node path={node[circle,fill,inner sep=1.3pt] at (\tikzlastnode)  {}}] (s2) at (7,0) {};$} for designated worlds. The planning agent, \Father, does not have access to this global state, since he considers it equally possible that the present is at \POne\ ($w_1$ and $w_2$ are indistinguishable to him). The internal initial state of the father is his associated local state $s_0^\Father = (\model, W)$:
\[ \raisebox{6mm}{$s_0^\Father =$ \quad} 
 \begin{tikzpicture}[every node/.style={auto},>=stealth]
\node[draw,circle,align=left,inner sep=2.3pt,label=below:{$w_1: \At(\Father,\Hom), \At(\Present,\POne)$},after node path={node[circle,fill,inner sep=1.3pt] at (\tikzlastnode)  {}}] (s1) at (0,0) {};
\node[draw,circle,inner sep=2.3pt,label=below:{$w_2: \At(\Father,\Hom), \At(\Present,\PTwo)$},after node path={node[circle,fill,inner sep=1.3pt] at (\tikzlastnode)  {}}] (s2) at (7,0) {};
\path[-] (s1) edge node[above] {\Father} (s2);
\end{tikzpicture}
\] 
We have $s \models \At(\Present,\PTwo)$ but $s^\Father \not\models \At(\Present,\PTwo)$: From the outside global perspective (the perspective of an omniscient external observer) it can initially be verified that the present is at \PTwo, but not from the perspective of the planning agent, \Father, himself.
\end{example}

In general, any belief state $B = \{ s_1,\dots,s_n \}$ (set of propositional states) can be equivalently represented as an epistemic state $((W,\sim,L),W)$ with $W = \{ w_1,\dots,w_n \}$, $\sim\ = W \times W$, $L(w_i) = s_i$ for all $i=1,\dots,n$.

With epistemic models we can also capture the distinction between run time and plan time uncertainty. When the father contemplates on the possible action sequences to execute in his local state $s^\Father$, the epistemic state representing the result of executing the action sequence \[ 
\Go(\Father,\Hom,\POne); \TryPickUp(\Father,\Present,\POne); \Go(\Father,\POne,\Hom)
\] will be
\[ \raisebox{6mm}{$s' =$ \quad} 
 \begin{tikzpicture}[every node/.style={auto},>=stealth]
\node[draw,circle,align=left,inner sep=2.3pt,label=below:{$w_1: \At(\Father,\Hom), \Has(\Father,\Present)$},after node path={node[circle,fill,inner sep=1.3pt] at (\tikzlastnode)  {}}] (s1) at (0,0) {};
\node[draw,circle,inner sep=2.3pt,label=below:{$w_2: \At(\Father,\Hom), \At(\Present,\PTwo)$},after node path={node[circle,fill,inner sep=1.3pt] at (\tikzlastnode)  {}}] (s2) at (7,0) {};
\end{tikzpicture}
\] 
There is still two designated worlds: the agent do not at plan time know which one it will end up in. But they are now not indistinguishable (there is no indistinguishability edge connecting them): at run time, when the plan is executed, the agent will be able to tell whether it is in $w_1$ or $w_2$. In fact we have e.g.\ $s' \models K_\Father \At(\Present,\PTwo) \vee K_\Father \neg \At(\Present,\PTwo)$: In $s'$, the father knows whether the present is at \PTwo\ or not. More generally, let $s = (\model,W_d)$ be any local state of an agent $i \in \agents$ and let $w_1,w_2 \in W_d$. The worlds $w_1$ and $w_2$ are called  \textbf{run time indistinguishable} (for agent $i$) if $w_1 \sim_i w_2$. Otherwise, they are \textbf{plan time indistinguishable}.

The generalisation from belief states (in the standard AI sense of sets of propositional states~\cite{russ.ea:arti}) to epistemic states (in the sense of epistemic logic) also allows us to represent planning tasks involving multiple agents. E.g.\ the father might want to make sure that his daughter doesn't get to know that he has a present for her (it is supposed to be a surprise). In this case, we could reformulate the goal as
\newcommand{\Daughter}{\textsf{Daughter}}
\begin{equation}\label{eq:extendedgoal} \small
   \phi_g =  \At(\Father,\Home) \land \Has(\Father,\Present) \land \Wrapped(\Present) \land \neg K_\Daughter \Has(\Father,\Present).
\end{equation}

\section{Dynamic epistemic logic}\label{sect:dynamicepistemiclogic}
Generalising from propositional states to epistemic states amounted to generalise from propositional valuations to multisets of such valuations with an added indistinguishability relation for each agent. We can now apply the exact same trick to generalise from propositional actions to epistemic actions. A propositional action is an event $\langle pre(a), post(a) \rangle$, so an epistemic action will be a multiset of events with an indistinguishability relation for each agent. This is exactly how epistemic actions are defined in dynamic epistemic logic (DEL) with postconditions~\cite{ditm.ea:sema}. Such structures are called \emph{event models} or \emph{action models}.
\begin{definition}
An \textbf{action model} on $P,\agents$ is $\mathcal{E} = (E,(\sim_i)_{i \in \agents},pre,post)$ where
\begin{itemize}
  \item The \textbf{domain} $E$ is a non-empty finite set of \textbf{events}. 
  \item $\sim_i\ \subseteq E \times E$ is an equivalence relation called the \textbf{indistinguishability relation} for agent $i \in \agents$.
  \item $pre: E \to \lang(P,\agents)$ assigns a \textbf{precondition} to each event.
  \item $post: E \to \lang(P,\agents)$ assigns a \textbf{postcondition} to each event. For all $e \in E$, $post(e)$ is a conjunction of literals over $P$ (including the special atom $\top$).\footnote{This definition of postconditions is slightly less general than the standard definition of postconditions in DEL~\cite{ditm.ea:sema}. Normally, $post(e)$ is a mapping from atomic propositions to formulas $\lang$, where $post(e)(p) = \phi$ means that $p$ after the event $e$ has occurred gets whatever truth value $\phi$ had before $e$ occurred. In~\cite{ditm.ea:sema} it is shown that any action model can be equivalently represented by one in which $post(e)(p) \in \{p, \top, \bot \}$ for all $e \in E$ and $p \in P$. This action model can potentially be exponentially larger. Such simplified (but potentially larger) action models can immediately be translated into the form defined here (see~\cite{bola.ea:epis} for details). The advantages of the conventions of this paper is to have a more clear connection to the conventions of classical planning and to simplify notation.}
\end{itemize}
For $E_d \subseteq E$, the pair $(\mathcal{E},E_d)$ is called an \textbf{epistemic action} (or simply an \textbf{action}), and the events in $E_d$ are called the \textbf{designated events}. An action is called \textbf{global} if $E_d = \{e \}$ for some event $e$, and we then often write $(\mathcal{E},e)$ instead of $(\mathcal{E}, \{e\})$. Similar to states, $(\mathcal{E}, E_d)$ is called a \textbf{local action} for agent $i$ when $E_d$ is closed under $\sim_i$. Given an action $a = (\mathcal{E},E_d)$, the \textbf{associated local action} of agent $i \in \agents$, denoted $a^i$, is $(\mathcal{E}, \{ f \in E \mid \text{$f \sim_i e$ for some $e \in E_d$} \})$. 
\end{definition}
Any propositional action $a = \langle pre(a), post(a) \rangle$ trivially \textbf{induces} an epistemic action $(\mathcal{E},\{e\})$ with $\mathcal{E} = (\{e\},\{(e,e)\},pre,post)$ where $pre(e) = pre(a)$ and $post(e) = post(a)$.
\begin{example}\label{ex:trypickupepistemic}
 We can now finally, using action models, give a satisfactory formal representation of the \TryPickUp\ action of Section~\ref{sect:beliefstates}, cf.\ (\ref{eq:trypickupgeneral}): 
\[ 
\begin{array}{l}
\TryPickUp(\agt,\obj,\from) = \\ \\
 \begin{tikzpicture}[every node/.style={auto},>=stealth]
\node[draw,circle,align=left,inner sep=2.3pt,label=below:{$\begin{array}{l} e_1: \langle \At(\agt,\from) \land \At(\obj,\from) \land \neg \Has(\agt,\obj), \\ \hspace{7mm} \Has(\agt,\obj) \land \neg \At(\obj,\from) \rangle \end{array}$},after node path={node[circle,fill,inner sep=1.3pt] at (\tikzlastnode)  {}}] (s1) at (0,0) {};
\node[draw,circle,inner sep=2.3pt,label=below:{$\begin{array}{l} e_2: \langle \At(\agt,\from) \land \neg \At(\obj,\from), \\ \hspace{7mm} \top \rangle \end{array}$},after node path={node[circle,fill,inner sep=1.3pt] at (\tikzlastnode)  {}}] (s2) at (8,0) {};
\end{tikzpicture}
\end{array}
\] 
We here use the same conventions for representing action models graphically as we did for epistemic models. 
Note that there is no indistinguishability edge for \Father: He will at run time observe whether the action is succesful ($e_1$) or not ($e_2$). But at plan time he does not know which one it will be, which is why both events are designated.
\end{example}
The state-transition function of DEL is called \emph{product update}, and is denoted by an infix $\otimes$ symbol. So instead of writing $\gamma(s,a)$, one writes $s \otimes a$. 
\begin{definition}
Let a state $s = (\model,W_d)$ and an action $a = (\mathcal{E},E_d)$ be given with $\model = (W,(\sim_i)_{i \in \agents}, L)$ and $\mathcal{E} = (E,(\sim_i)_{i \in \agents},pre, post)$. Then the \textbf{product update} of $s$ with $a$ is $s \otimes a = ((W',(\sim'_i)_{i \in \agents},L'),W'_d)$ where
\begin{itemize}
  \item $W' = \{ (w,e) \in W \times E \mid (\model,w) \models pre(e) \}$
  \item $\sim'_i \ = \{((w,e), (w',e')) \in W' \times W' \mid w \sim_i w' \text{ and } e \sim_i e' \}$
  \item $L'(p) =  (L(p) - \{ p \mid \neg p \text{ is a negative literal of } post(e)) \cup
    \{ p \mid p \text{ is a positive literal of } post(e) \}$
  \item $W'_d = \{ (w,e) \in W' \mid w \in W_d \text{ and } e \in E_d \}$.
\end{itemize}
We say that an action $(\mathcal{E},E_d)$ is \textbf{applicable} in a state $(\model,W_d)$ if for all $w \in W_d$ there is an event $e \in E_d$ such that $(\model,w) \models pre(e)$.
\end{definition}
Note that this is the obvious generalisation to the epistemic setting of the state-transition function for belief states and sets of events given in (\ref{form:gammabeliefstates}). In particular, if $s'$ is the epistemic state induced by a belief state $s$, and $a'$ is  the action model induced by a propositional action $a$, then $s' \otimes a'$ is the epistemic state induced by $\gamma(s,a)$.

Let $i \in \agents$ be an agent, let $(\model,W_d)$ be a local state for $i$ and let $(\mathcal{E},E_d)$ be a local action for $i$. Then $(\model,W_d)$ and $(\mathcal{E},E_d)$ represent agent $i$'s view on a particular state and action. According to the definition above, the action $(\mathcal{E},E_d)$ is then \emph{applicable} in the state $(\model,W_d)$ if, for any of the worlds agent $i$ considers possible (any world $w$ in $W_d$), the action specifies at least one applicable event that agent $i$ considers possible (an event $e$ in $E_d$ with $(\model,w) \models pre(e)$). In other words, the action is applicable from the perspective of agent $i$. 

\section{Epistemic planning tasks}\label{sect:epistemicplanningtasks}
We now have all the necessary ingredients for defining planning tasks in epistemic planning based on DEL. We simply define these as for propositional planning tasks, except we replace propositional actions by epistemic actions, propositional states by epistemic states and propositional goal formulas by epistemic goal formulas.
\begin{definition}
An \textbf{epistemic planning task} on $P,\agents$ is $(A,s_0,\phi_g)$ where $A$ is a finite set of epistemic actions on $P,\agents$, $s_0$ is an epistemic state on $P,\agents$, and $\phi_g \in \lang(P,\agents)$. We say that $(A,s_0,\phi_g)$ is an epistemic planning task \textbf{for agent $i \in \agents$} if $s_0$ and all $a\in A$ are local for $i$. A planning task $(A,s_0,\phi_g)$ is called \textbf{global} if $s_0$ is global. Given any planning task $\Pi = (A,s_0,\phi_g)$, the \textbf{associated local planning task} of agent $i$, denoted $\Pi^i$, is $(\{a^i \mid a \in A \}, s_0^i, \phi_g)$.\footnote{Given a planning task $\Pi$, the planning task $\Pi^i$ is agent $i$'s local perspective on that planning task. E.g.\ if the present is at \POne, then from an outside perspective, the planning task has an initial state where only the world satisfying $\At(\Present,\POne)$ is designated, but agent \Father\ is still forced to do planning based on the planning task $\Pi^\Father$ where both the $\At(\Present,\POne)$-world and the $\At(\Present,\PTwo)$-world are designated, since the father doesn't know which of the two is the actual initial state.}
\end{definition}
An epistemic planning task $(A,s_0,\phi_g)$ \textbf{induces} a classical planning task $((S,A,\gamma),s_0,S_g)$ in a similar manner to propositional planning tasks:
\begin{itemize}
  \item $S = \{ s_0 \otimes a_1 \otimes \cdots \otimes a_n \mid n \in \mathbb{N} \text{ and } a_1,\dots,a_n \in A\}$ 
  \item $\gamma(s,a) = \begin{cases} 
    s \otimes a &\text{if $a$ is applicable in $s$} \\
    \text{undefined} &\text{otherwise}
    \end{cases}$
  \item $S_g = \{ s \in S \mid s \models \phi_g \}$.
\end{itemize}
As for propositional planning tasks, we can then define a \textbf{solution} to an epistemic planning task $(A,s_0, \phi_g)$ to be a solution to the induced classical planning task. This is equivalent to the existence of a sequence of actions $a_1; \cdots; a_n$ from $A$ such that each $a_i$ is applicable in $s_0 \otimes a_1 \otimes \cdots \otimes a_{i-1}$ ($i \leq n$) and $s_0 \otimes a_1 \otimes \cdots \otimes a_n \models \phi_g$.
\begin{example}\label{ex:epistemicplanning}
Let us provide a full formalisation of the birthday present example as an epistemic planning task. The epistemic action \TryPickUp\ was presented in Example~\ref{ex:trypickupepistemic}. The actions $\Go(\agt,\from,\myto)$ and $\Wrap(\agt,\obj)$ are simply the ones induced by their propositional counterparts. The initial state is the state $s_0^\Father$ of Example~\ref{ex:actionmodels}. The goal formula is $\phi_g = \At(\Father,\Home) \land \Has(\Father,\Present) \land \Wrapped(\Present)$. Clearly the resulting planning task is a planning task for agent \Father. We now get
\[ 
  \begin{array}{l}
    s_1 = s_0^\Father \otimes \Go(\Father,\Hom,\POne)
           = \raisebox{-5mm}{\begin{tikzpicture}[every node/.style={auto},>=stealth]
\node[draw,circle,align=left,inner sep=2.3pt,label=below:{$\begin{array}{l} \At(\Father,\POne), \\ \At(\Present,\POne) \end{array}$},after node path={node[circle,fill,inner sep=1.3pt] at (\tikzlastnode)  {}}] (s1) at (0,0) {};
\node[draw,circle,inner sep=2.3pt,label=below:{$\begin{array}{l} \At(\Father,\POne), \\ \At(\Present,\PTwo) \end{array}$},after node path={node[circle,fill,inner sep=1.3pt] at (\tikzlastnode)  {}}] (s2) at (3.5,0) {};
\path[-] (s1) edge node[above] {\Father} (s2);
\end{tikzpicture}} \\ \\
   s_2 = s_1 \otimes \TryPickUp(\Father,\Present,\POne) 
          =  \raisebox{-5mm}{  
          \begin{tikzpicture}[every node/.style={auto},>=stealth]
\node[draw,circle,align=left,inner sep=2.3pt,label=below:{$\begin{array}{l} \At(\Father,\POne), \\ \Has(\Father,\Present) \end{array}$},after node path={node[circle,fill,inner sep=1.3pt] at (\tikzlastnode)  {}}] (s1) at (0,0) {};
\node[draw,circle,inner sep=2.3pt,label=below:{$\begin{array}{l} \At(\Father,\POne), \\ \At(\Present,\PTwo) \end{array}$},after node path={node[circle,fill,inner sep=1.3pt] at (\tikzlastnode)  {}}] (s2) at (3.5,0) {};
\end{tikzpicture}
}
\end{array}
\]
Note that the indistinguishability link between the two worlds is being cut when going from $s_1$ to $s_2$, since the two events of \TryPickUp\ are distinguishable. We further get
\[ \small
  \begin{array}{l}
    s_3 = s_2 \otimes \Go(\Father,\POne,\PTwo)
           = \raisebox{-5mm}{\begin{tikzpicture}[every node/.style={auto},>=stealth]
\node[draw,circle,align=left,inner sep=2.3pt,label=below:{$\begin{array}{l} \At(\Father,\PTwo), \\ \Has(\Father,\Present) \end{array}$},after node path={node[circle,fill,inner sep=1.3pt] at (\tikzlastnode)  {}}] (s1) at (0,0) {};
\node[draw,circle,inner sep=2.3pt,label=below:{$\begin{array}{l} \At(\Father,\PTwo), \\ \At(\Present,\PTwo) \end{array}$},after node path={node[circle,fill,inner sep=1.3pt] at (\tikzlastnode)  {}}] (s2) at (3.5,0) {};
\end{tikzpicture}} \\ \\
   s_4 = s_3 \otimes \TryPickUp(\Father,\Present,\PTwo) 
          =  \raisebox{-5mm}{  
          \begin{tikzpicture}[every node/.style={auto},>=stealth]
\node[draw,circle,align=left,inner sep=2.3pt,label=below:{$\begin{array}{l} \At(\Father,\PTwo), \\ \Has(\Father,\Present) \end{array}$},after node path={node[circle,fill,inner sep=1.3pt] at (\tikzlastnode)  {}}] (s1) at (0,0) {};
\node[draw,circle,inner sep=2.3pt,label=below:{$\begin{array}{l} \At(\Father,\PTwo), \\ \Has(\Father,\Present) \end{array}$},after node path={node[circle,fill,inner sep=1.3pt] at (\tikzlastnode)  {}}] (s2) at (3.5,0) {}; 
\end{tikzpicture} 
} \\ \\
  s_6 = s_4 \otimes \Go(\Father,\PTwo,\Hom) \otimes \Wrap(\Father,\Present) 
          = \hspace{-3mm}  \raisebox{-8mm}{  
          \begin{tikzpicture}[every node/.style={auto},>=stealth]
\node[draw,circle,align=left,inner sep=2.3pt,label=below:{$\begin{array}{l} \At(\Father,\Home), \\ \Has(\Father,\Present) \\ \Wrapped(\Present) \end{array}$},after node path={node[circle,fill,inner sep=1.3pt] at (\tikzlastnode)  {}}] (s1) at (0,0) {};
\node[draw,circle,inner sep=2.3pt,label=below:{$\begin{array}{l} \At(\Father,\Home), \\ \Has(\Father,\Present) \\ \Wrapped(\Present) \end{array}$},after node path={node[circle,fill,inner sep=1.3pt] at (\tikzlastnode)  {}}] (s2) at (3.5,0) {};
\end{tikzpicture}
}
\end{array}
\]
Since the two worlds of $s_6$ have identical labels, we can take the \emph{bisimulation contraction} which will preserve only one of them.\footnote{For a definition of bisimulation between epistemic models with multiple designated points, see~\cite{bola.ea:epis}. For a definition of bisimulation on models of modal logic in general, and a definition of bisimulation contraction, see~\cite{blac.ea:moda}. Bisimulation contractions are an indispensible tool in practice when working with product updates in DEL, since sequences of updates otherwise tend to produce very large model.} We clearly have $s_5 \models \phi_g$. Hence the action sequence (\ref{eq:planmodified}) above is a solution to the epistemic planning task. We could actually replace the $\TryPickUp(\Father,\Present,\PTwo)$ action in this plan with the epistemic action induced by the original non-conditional $\PickUp(\Father,\Present,\PTwo)$ action. This is because the father will know that the present is at \PTwo\ when he gets there (since it wasn't at \POne).  
\end{example}

\section{Conditional epistemic planning}\label{sect:conditionalepistemicplanning}
The plan found in the previous example, Example~\ref{ex:epistemicplanning}, is clearly not always optimal. If the father gets the present already at \POne, there is no need to go to \PTwo\ as well. So in this case the father will be able to reach the goal in 4 instead of 6 steps. But the father will of course not know this until run time (after having attempted to pick up the present at \POne). A sequential plan (sequence of actions) can not capture this, since it has a fixed number of actions, independent of the action outcomes. So we need to move to \emph{conditional plans} (not to be confused with \emph{conditional actions} that we already have). A conditional plan for the current planning task could e.g.\ be formulated as the following program
\[
  \begin{array}{l}
   \Go(\Father,\Hom,\POne); \TryPickUp(\Father,\Present,\POne); \\
   \textbf{if } K_\Father \Has(\Father,\Present) \textbf{ then } \Go(\Father,\POne,\Hom); \Wrap(\Father,\Present) \textbf{ else } \\
   \Go(\Father,\POne,\PTwo); \TryPickUp(\Father,\Present,\PTwo); \Go(\Father,\PTwo,\Hom); \\
   \Wrap(\Father,\Present)
  \end{array}
\]
Such programs for epistemic planning tasks are formally defined and explored in~\cite{ande.ea:cond} (single-agent case only).\footnote{Note that the if-condition of the if-then-else construct is a $K$-formula expressing knowledge of the planning agent (\Father). All such if-conditions are required to be $K$-formulas of the planning agent, as an agent can only make a choice based on what he knows. Otherwise we could simply construct a conditional plan like $\textbf{if } \At(\Present,\POne) \textbf{ then } \Go(\Father,\Home,\POne) \textbf{ else } \Go(\Father,\Home,\PTwo)$. However, in this case the agent would not know how to settle the truth-value of the if-condition $\At(\Present,\POne)$ in the initial state, and would hence not know whether to execute $\Go(\Father,\Home,\POne)$ or $\Go(\Father,\Home,\PTwo)$.} Alternatively, such programs can be seen as abbreviations of PDL programs, e.g.\ the program $\textbf{if } \phi \text{ then } \pi_1 \textbf{ else } \pi_2$ can be seen as shorthand for the PDL program $(\phi?; \pi_1) \cup (\neg \phi?; \pi_2)$. In \cite{eijck2014dynamic}, dynamic epistemic logic with postconditions and PDL constructs is studied (however not in a planning context). Programs like the one above are in~\cite{fagi.ea:reas} called \emph{knowledge-based programs} and are there studied in an alternative logical setting. Knowledge-based programs are studied in a planning context in~\cite{lang2013knowledge}. 

An alternative to conditional plans as programs is \emph{policies}. A policy is a mapping from states into actions, that is, for each state the policy specifies the action chosen in that state (a policy is also sometimes called a \emph{strategy} or a \emph{protocol}, depending on the area). Since implemented planning systems often generate conditional plans via an \textsc{And-Or} graph search in the state space, a policy comes as a more direct output of these algorithms than a program (e.g.\ one does not have to synthesise appropriate if-conditions). From now on we will only consider conditional plans as policies, not programs.
\begin{definition}
  Let $\Pi = (A,s_0,\phi_g)$ be a planning task and $i \in \agents$ be an agent. An $i$-policy $\pi$ for $\Pi$ is a partial mapping $\pi: \sglobal \hookrightarrow A$ from global states into actions such that 
  \begin{enumerate}
    \item[(1)] If $\pi(s) = a$ then $a$ is applicable in $s^i$. 
    \item[(2)] If $s^i = t^i$ then $\pi(s) = \pi(t)$.
  \end{enumerate}
\end{definition}
An $i$-policy is a policy from the perspective of agent $i$. Condition (1) ensures that in such a policy, agent $i$ always knows that the next action to be performed is applicable (\emph{knowledge of preconditions}). Condition (2) is a \emph{uniformity condition}: If two global states are indistinguishable to agent $i$, agent $i$ has to make the same choice in both.
\begin{definition}
An \textbf{execution} of a policy $\pi$ from a global state $s_0$ is a maximal (finite or infinite) sequence of alternating global states and actions $(s_0,a_1,s_1,a_2,s_2,\dots)$ such that for all $m\geq0$,
\begin{enumerate}
  \item[(1)] $\pi(s_m) = a_{m+1}$, and
  \item[(2)] $s_{m+1} \in \globals(s_m \otimes a_{m+1})$.
\end{enumerate}
An execution is called \textbf{successful} for a global planning task $\Pi = (A,s_0,\phi_g)$ if it is a finite execution $(s_0,a_1,s_1,\dots,a_n,s_n)$ such that $s_n \models \phi_g$.
\end{definition}
\begin{definition}
A policy $\pi$ for a planning task $\Pi = (A,s_0,\phi_g)$ is called a \textbf{solution} to $\Pi$ if $\globals(s_0) \subseteq \dom(\pi)$ and for each $s \in \dom(\pi)$, any execution of $\pi$ from $s$ is successful for $\Pi$.\footnote{Note that a solution policy has to lead to successful executions for all global states in $s_0$ (and for all global states along all possible executions of the policy). Such policies are often called \emph{strong policies}, because the requirement is that they are guaranteed to succeed under alle circumstances (considered possible by the planning agent). In conditional planning, one often also considers weak solutions (\emph{weak policies}) that have a possibility of succeeding, but might not always succeed. A weak solution to the birthday present task would be to go to \POne, attempt picking up the present there, go home, and then wrap the present if one has it. This plan can clearly fail if the present is at \PTwo. In this paper we restrict attention to strong solutions.}
\end{definition}

\begin{example}
In the birthday present task we can now specify distinct actions to be performed in the two distinct global states of the state $s_2$ of Example~\ref{ex:epistemicplanning} (the state after the father has attempted to pickup the present at \POne):
\[
   \begin{array}{l}
     \pi( \raisebox{-6mm}{\begin{tikzpicture}[every node/.style={auto},>=stealth]
\node[draw,circle,align=left,inner sep=2.3pt,label=below:{$\begin{array}{l} \At(\Father,\POne), \\ \Has(\Father,\Present) \end{array}$},after node path={node[circle,fill,inner sep=1.3pt] at (\tikzlastnode)  {}}] (s1) at (0,0) {};
\node[draw,fill,circle,inner sep=1.3pt,label=below:{$\begin{array}{l} \At(\Father,\POne), \\ \At(\Present,\PTwo) \end{array}$}]  (s2) at (3.5,0) {};
\end{tikzpicture}}) = \Go(\Father,\POne,\Home) \\ \\
 \pi( \raisebox{-6mm}{\begin{tikzpicture}[every node/.style={auto},>=stealth]
\node[draw,fill, circle,align=left,inner sep=1.3pt,label=below:{$\begin{array}{l} \At(\Father,\POne), \\ \Has(\Father,\Present) \end{array}$}] (s1) at (0,0) {};
\node[draw,circle,inner sep=2.3pt,label=below:{$\begin{array}{l} \At(\Father,\POne), \\ \At(\Present,\PTwo) \end{array}$},after node path={node[circle,fill,inner sep=1.3pt] at (\tikzlastnode)  {}}]  (s2) at (3.5,0) {};
\end{tikzpicture}}) = \Go(\Father,\POne,\PTwo).
  \end{array}
\]
This partial policy can easily be extended into a full solution policy to the planning task. 
\end{example}
\newcommand{\Ask}{\textsf{Ask}}
\newcommand{\Employee}{\textsf{Employee}}
\begin{example}
Consider extending the birthday present example with a post office employee, \Employee, working at \POne. We can think of this as a new agent, so now $\agents = \{ \Father, \Employee \}$. It might be possible to call the employee from home to ask whether the present is at the post office. We can assume the employee knows whether the present is at his or her post office, so the initial state would be the same as in Example~\ref{ex:epistemicplanning} ($s_0^\Father$), except there is now reflexive loops for \Employee\ at both worlds. We can now represent a general action of an agent $i$ calling an agent $j$ to ask whether a formula $\phi$ is true as the following action, where $i,j \in \agents$ and $\phi \in \lang$:
\[ 
\begin{array}{l}
\Ask(i,j,\phi) = 
 \raisebox{-4mm}{\begin{tikzpicture}[every node/.style={auto},>=stealth]
\node[draw,circle,align=left,inner sep=2.3pt,label=below:{$\langle K_j \phi, \top \rangle$},after node path={node[circle,fill,inner sep=1.3pt] at (\tikzlastnode)  {}}] (s1) at (0,0) {};
\node[draw,circle,inner sep=2.3pt,label=below:{$\langle K_j \neg \phi, \top \rangle$},after node path={node[circle,fill,inner sep=1.3pt] at (\tikzlastnode)  {}}] (s2) at (2,0) {};
\node[draw,circle,inner sep=2.3pt,label=below:{$\langle \neg K_j \phi \land \neg K_j \neg \phi, \top \rangle$},after node path={node[circle,fill,inner sep=1.3pt] at (\tikzlastnode)  {}}] (s2) at (4.9,0) {};
\end{tikzpicture}}
\end{array}
\] 
The action model above corresponds to agent $j$ giving a sincere answer to the question ``is $\phi$ true?''. Agent $j$ saying ``yes'' to the question is represented by the event $e_1$, that is, it corresponds to an announcement that agent $j$ knows $\phi$. Event $e_2$ corresponds to saying ``no'', and $e_3$ corresponds to saying ``I don't know''. We now get
\[
     s_0^\Father \otimes \Ask(\Father,\Employee,\At(\Present,\POne))
           = \raisebox{-5mm}{\begin{tikzpicture}[every node/.style={auto},>=stealth]
\node[draw,circle,align=left,inner sep=2.3pt,label=below:{$\begin{array}{l} \At(\Father,\Hom), \\ \At(\Present,\POne) \end{array}$},after node path={node[circle,fill,inner sep=1.3pt] at (\tikzlastnode)  {}}] (s1) at (0,0) {};
\node[draw,circle,inner sep=2.3pt,label=below:{$\begin{array}{l} \At(\Father,\Home), \\ \At(\Present,\PTwo) \end{array}$},after node path={node[circle,fill,inner sep=1.3pt] at (\tikzlastnode)  {}}] (s2) at (3.5,0) {};
\end{tikzpicture}}
\]
This is because in $w_1$ of $s_0^\Father$ we have that $K_\Employee \At(\Present,\POne)$ holds and in $w_2$ of $s_0^\Father$ we have $K_\Employee \neg \At(\Present,\POne)$, and the action model for $\Ask(\Father,\Employee,\At(\Present,\POne))$ is clearly seen to cut the link between worlds satisfying $K_\Employee \neg \At(\Present,\POne)$ and worlds satisfying $K_\Employee \neg \At(\Present,\POne)$ (since there are two distinguishable events with these formulas as preconditions). Hence the planning agent, \Father, can conclude
\[
  s_0 \otimes \Ask(\Father,\Employee,\At(\Present,\POne)) \models K_\Father \At(\Present,\POne) \vee K_\Father \At(\Present, \PTwo)
\]
and hence that he can first call to ask the post office employee whether she has the present, and then next he will know where to go. 
 \end{example}
 
\section{Public and private actions}
 
\newcommand{\EmployeeTwo}{\textsf{Employee2}} 
The \Ask\ action above is a bit simplified. It is \textbf{publicly observable}:  the indistinguishability relation is the identity. When there is only two agents in the scenario this is acceptable, since they can only call each other, but if there were more agents present, these would probably not observe the phone call taking place. Consider adding a third agent, \EmployeeTwo, working at \PostOfficeTwo. If \Father\ calls \Employee, \EmployeeTwo\ will not know, and might not even consider it possible that the phone call could have taken place. To model this we have to go beyond equivalence classes in epistemic models and action models, since now it becomes possible to have false beliefs, e.g.\ a false belief that no phone call took place (or a false belief that \Father\ does not yet know $\At(\Present,\PTwo)$). In this case, we could model the private phone call of agent $i$ to agent $j$ asking about $\phi$ as follows:
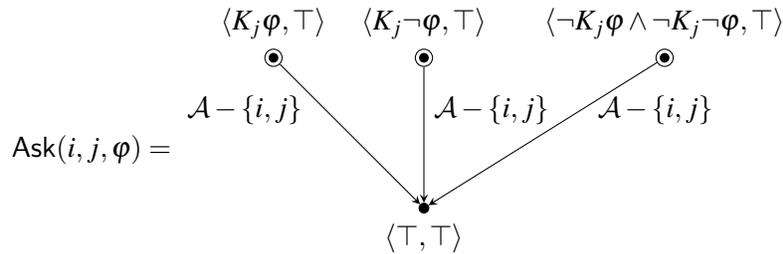
\[ 
\begin{array}{l}
\Ask(i,j,\phi) = 
 \raisebox{-14mm}{\begin{tikzpicture}[every node/.style={auto},>=stealth]
\node[draw,circle,align=left,inner sep=2.3pt,label=above:{$\langle K_j \phi, \top \rangle$},after node path={node[circle,fill,inner sep=1.3pt] at (\tikzlastnode)  {}}] (s1) at (0,0) {};
\node[draw,circle,inner sep=2.3pt,label=above:{$\langle K_j \neg \phi, \top \rangle$},after node path={node[circle,fill,inner sep=1.3pt] at (\tikzlastnode)  {}}] (s2) at (2,0) {};
\node[draw,circle,inner sep=2.3pt,label=above:{$\langle \neg K_j \phi \land \neg K_j \neg \phi, \top \rangle$},after node path={node[circle,fill,inner sep=1.3pt] at (\tikzlastnode)  {}}] (s3) at (5.2,0) {};
\node[draw,circle,fill,inner sep=1.3pt,label=below:{$\langle \top, \top \rangle$}] (s4) at (2,-2) {};
\path[->] (s1) edge node[above,xshift=-14mm] {$\agents - \{i,j\}$} (s4);
\path[->] (s2) edge node[above,xshift=9mm] {$\agents - \{i,j\}$} (s4);
\path[->] (s3) edge node[above,xshift=15mm] {$\agents - \{i,j\}$} (s4);
\end{tikzpicture}}
\end{array}
\] 
This models that agent $i$ and $j$ know the phone call takes place (and know the outcome of the phone call), but all other agents think that nothing has happened (the \emph{skip} event $\langle \top, \top \rangle$). We can even model that some subset $\mathcal{B} \subseteq \agents$ of agents hear the question being asked, but not the answer (they are together with $i$, but cannot hear what is being said in the other end):
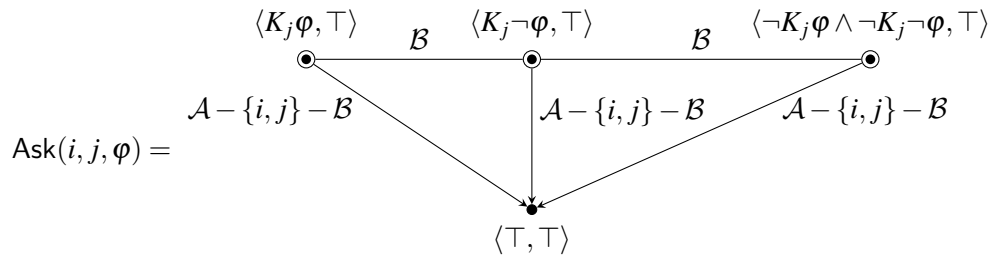
\[ 
\begin{array}{l}
\Ask(i,j,\phi) = 
 \raisebox{-14mm}{\begin{tikzpicture}[every node/.style={auto},>=stealth]
\node[draw,circle,align=left,inner sep=2.3pt,label=above:{$\langle K_j \phi, \top \rangle$},after node path={node[circle,fill,inner sep=1.3pt] at (\tikzlastnode)  {}}] (s1) at (-1,0) {};
\node[draw,circle,inner sep=2.3pt,label=above:{$\langle K_j \neg \phi, \top \rangle$},after node path={node[circle,fill,inner sep=1.3pt] at (\tikzlastnode)  {}}] (s2) at (2,0) {};
\node[draw,circle,inner sep=2.3pt,label=above:{$\langle \neg K_j \phi \land \neg K_j \neg \phi, \top \rangle$},after node path={node[circle,fill,inner sep=1.3pt] at (\tikzlastnode)  {}}] (s3) at (6.5,0) {};
\node[draw,circle,fill,inner sep=1.3pt,label=below:{$\langle \top, \top \rangle$}] (s4) at (2,-2) {};
\path[->] (s1) edge node[above,xshift=-20mm] {$\agents - \{i,j\} - \mathcal{B}$} (s4);
\path[->] (s2) edge node[above,xshift=12mm] {$\agents - \{i,j\} - \mathcal{B} $} (s4);
\path[->] (s3) edge node[above,xshift=22mm] {$\agents - \{i,j\} - \mathcal{B}$} (s4);
\path[-] (s1) edge node[above] {$\mathcal{B}$} (s2);
\path[-] (s2) edge node[above] {$\mathcal{B}$} (s3);
\end{tikzpicture}}
\end{array}
\] 

These examples illustrate that designing action models to represent actions under multi-agent partial observability is far from a trivial matter. Some recent papers try to propose different ways of encoding observability information into a logical language, so that the action models themselves can become simpler, more uniform, and automatically induced from the underlying observability information~\cite{bolander2014seeing,bolander2015announcements,herzig2015poor}. However, there are still many unsolved problems in this area.

The possibility of representing \emph{private actions} (only observed by a subset of agents, whereas the other agents believe nothing happens) is also what allows us to model the full version of the birthday present example, where the daughter does not get to know that the father has the present. We can for instance model that only agents who are \emph{copresent} (in the same location) can observe actions taking place in that location (see \cite{brenner2009continual} for a discussion and treatment of copresence in an epistemic planning setting). For instance, the \Wrap\ action could be reformulated as follows:
\newcommand{\loc}{\textit{loc}}
\[
\begin{array}{l}
\Wrap(\agt,\obj,\loc) = 
 \raisebox{-11mm}{\begin{tikzpicture}[every node/.style={auto},>=stealth]
\node[draw,circle,align=left,inner sep=2.3pt,label=below:{$\begin{array}{l} \langle \At(\agt,\loc) \land \Has(\agt,\obj) \land \neg \Wrapped(\obj), \\ \Wrapped(\obj) \rangle \end{array}$},after node path={node[circle,fill,inner sep=1.3pt] at (\tikzlastnode)  {}}] (s1) at (0,0) {};
\node[draw,circle,inner sep=2.3pt,label=below:{$\langle \top, \top \rangle$},after node path={node[circle,fill,inner sep=1.3pt] at (\tikzlastnode)  {}}] (s2) at (6,0) {};
\path[->] (s1) edge node[above] {$\{ i: \neg \At(i,\loc) \}_{i \in \agents}$} (s2);
\end{tikzpicture}}
\end{array}
\]
We here use a new type of action model that we call an \textbf{edge-conditioned action model}~\cite{bolander2014seeing}. The meaning of the label on the edge is that there is an edge for agent $i$ here if $\neg \At(i,\loc)$ is true (agent $i$ is not present in the location where the present is being wrapped). Such edge-conditioned action models are a variant of \emph{generalised arrow updates}~\cite{kooi2011generalized}. 

With the wrapping action presented as above, the father will e.g.\ be able to figure out that if the daughter is at home ($\At(\Daughter,\Home)$ is true), then only the plan where he executes \Wrap\ at the post office will lead to $\Wrapped(\Present) \wedge \neg K_\Daughter \Has(\Father,\Present)$ (recall the state-transition system in Figure~\ref{figu:stripsactions} that shows that the present can either be wrapped in the post office or at home). 
 
\section{Multi-agent planning}\label{sect:multiagent}
So far we have only considered the case of a single agent planning in the presence of other agents. The formal framework of course also allows to represent real multi-agent planning tasks. However, for true multi-agent planning a lot of additional design choices have to be made. Are the agents collaborating or competing? Are there \emph{coalitions} of agents trying to achieve a joint goal, perhaps against the agents outside the coalition? Can the agents communicate and coordinate arbitrarily when coming up with a plan, and will they commit to such a joint plan? The papers~\cite{engesser2017cooperative,bolander2016better} look at \emph{implicitly coordinated plans}, where agents have a joint goal but are not allowed to coordinate or negotiate in advance. All coordination should happen during plan executions as a result of announcements and observing the actions of others. 

In the birthday present example it could e.g.\ be that \Father\ calls \Employee\ to announce the goal $\Has(\Father,\Present)$. Assuming \Employee\ is altruistic and adopts this as his or her own goal, \Father\ and \Employee\ are now engaged in planning with implicit coordination towards the joint goal $\Has(\Father,\Present)$. Assume the present is at \PostOfficeTwo. Since \Employee\ knows that \Father\ doesn't know the location of the present, and that he cannot get the present unless he knows, she will choose to announce that the present is at \PostOfficeTwo. This is a case of implicit coordination: The father doesn't ask the employee about the location of the present, but the employee knows about the goal and can still plan to inform him, in order to allow him to plan the rest of the required actions. These ideas are explored further in~\cite{engesser2017cooperative,bolander2016better}.

\section{(Un)decidability and complexity}\label{sect:complexity}
One of the most studied problems in epistemic planning based on DEL is the complexity of the plan existence problems. The \textbf{plan existence problem} for a class of planning tasks $X$ is the following decision problem: Given a planning task $\Pi \in X$, does there exist a solution to $\Pi$? So far only the complexity of deciding whether a sequential plan exists have been studied. For general epistemic planning tasks the problem is undecidable already with two agents, no common knowledge, and even without postconditions (that is, purely epistemic planning without ontic change)~\cite{auch.ea:unde}. This has lead to a quest for finding decidable, but still reasonably expressive, fragments of epistemic planning. The first result along these lines proved that epistemic planning with propositional preconditions (that is, no epistemic formulas in the preconditions) is decidable~\cite{yu2013multi}. An upper bound on complexity is in that paper shown to be $(n+1)$-\textsc{ExpTime} for planning tasks in which the goal formula has modal depth $n$. In~\cite{charrier2016impact} it is shown that when the preconditions are propositional \emph{and} there are no postconditions, then the plan existence problem is \textsc{PSpace}-complete (for arbitrary goal formulas). If restricting further to certain types of actions, like private and public announcements, complexity of plan existence  goes further down to \textsc{NP}-complete~\cite{bolander2015complexity}.

\section{Alternative approaches to epistemic planning}\label{sect:alternative}
Epistemic planning based on DEL takes a \emph{semantic approach}, where states are represented as semantical objects, epistemic states. It is also possible to take a \emph{syntactic approach}, where states are represented as knowledge-bases, sets of formulas known to be true. Interestingly, STRIPS and propositional planning are semantic and syntactic at the same time, since in propositional logic a semantic state is just a set of formulas (a set of true atomic propositions). This means that when generalising from STRIPS or propositional planning to epistemic planning, it is not immediately obvious whether a semantic or syntactic approach would be most appropriate. Both approaches have their strength and weaknesses. 

Epistemic planning based on DEL also takes the approach of insisting on succinct action representations via action models. We argued in favour of this approach already very early in the paper by mentioning the weaknesses of attempting to represent planning tasks explicitly as state-transition systems. In fact, the state-transition systems induced by epistemic planning tasks will often be infinite, so even if choosing a state-transition system based approach, one will need a way to represent them finitely in order to be able to do planning based on them~\cite{jamroga2007constructive}. From the perspective of classical planning, the most appropriate approach to representing state-transition systems compactly seems to be representing the actions themselves in a compact way (using some kind of action models or action schemas).

Based on the above, we can  distinguish approaches to epistemic planning along two dimensions:
\begin{itemize}
  \item \textbf{Semantic approaches} versus \textbf{syntactic approaches}.
  \item \textbf{Action model based approaches} versus \textbf{state-transition system based approaches}.
\end{itemize}
The approach of this paper is, as mentioned, semantic and action model based. Syntactic approaches to epistemic planning can be found in e.g.\ the (single-agent) PKS planner \cite{petr.ea:pks}, the (multi-agent) planning framework of \cite{muise2015planning} and the compilation approach of \cite{kominis2014beliefs}, translating a restricted fragment of epistemic planning into classical planning. The state-transition system based approach is found in a number of papers in temporal epistemic logic, e.g.~\cite{hoek.ea:trac,jamroga2007constructive}. In these papers, the planning domains are represented by a type of epistemic state-transition system called a \emph{concurrent epistemic game structure} (CEGS). 
Having an explicitly represented state-transition system like a CEGS tends to make it easier to define complex notions of multi-agent plans (strategies). On the other hand, by working directly with state-transition systems, some of the important problems of epistemic planning are being silently bypassed. These problems include compact representations of the state-transition system, and efficient heuristics for avoiding to build the entire state-transition system when planning. They also include problems of how to provide a general approach to multi-agent observability and problems of concurrent composition of actions of multiple agents. In CEGS, transitions represent \emph{joint actions}, one action per agent. The problem of how to make a parallel composition of the actions of the individual agents has to be solved before one can even build the CEGS for a given planning domain. In epistemic planning based on DEL, this problem has to be solved at the level of action models: How do we make a parallel composition of two action models in case they are not independent (e.g.\ two agents trying to open a door at the same time, but from opposite sides). Epistemic planning tasks can be seen as inducing e.g.\ forests of epistemic temporal logic~\cite{bent.ea:merg} or CEGSs. The exact relations have not been explored yet, but providing a link between the action model based approaches and the state-transition system based approaches could be very valuable and provide a possibility of getting the best of both worlds. 

Similarly, providing links and connections between syntactic and semantic approaches seem to be potentially very valuable. In a semantic approach, one is essentially modelling \emph{ignorance}: the more ignorance, the bigger the state. In a syntactic approach, one is essentially modelling \emph{knowledge}: the more knowledge, the bigger the state. Knowledge and ignorance are each others duals, and hence the semantic and the syntactic approach are also each others duals. It seems that when humans are planning, we are sometimes using a more semantic approach, and sometimes a more syntactic one, depending on the planning task at hand. It would hence also be very interesting to see whether planning frameworks could be developed that would employ or combine both approaches.

\bibliographystyle{eptcs}
\bibliography{litt}
\end{document}